# HouYi：An open-source large language model specially designed for renewable energy and carbon neutrality field

Mingliang Bai[1], Zhihao Zhou[1], Ruidong Wang[1], Yusheng Yang[2],

Zizhen Qin[1], Yunxiao Chen[1], Chunjin Mu[1], Jinfu Liu[1]*, Daren Yu[1]

1. Harbin Institute of Technology, 150001, Harbin, Heilongjiang, China
2. Peking University Shenzhen Graduate School, 518000, Shenzhen, Guangdong, China

**Abstract:** Renewable energy is important for achieving carbon neutrality goal. With the great success of Large Language Models (LLMs) like ChatGPT in automatic content generation, LLMs are playing an increasingly important role. However, there has not been a specially designed LLM for renewable energy. Meanwhile, there has not been any dataset of renewable energy for training LLMs. Therefore, this paper published the first open-source Renewable Energy Academic Paper (REAP) dataset for non-commercial LLM research of renewable energy. REAP dataset is collected through searching the title and abstract of 1,168, 970 academic literatures from Web of Science. Based on REAP dataset, HouYi ("后羿" in Chinese) model, the first LLM for renewable energy, is developed through finetuning general LLMs. HouYi demonstrated powerful academic paper paragraph generation ability in renewable energy field. Experiments show that its ability to generate academic papers on renewable energy is comparable to ChatGPT, slightly outperforms Claude, ERNIE Bot and SparkDesk, and significantly outperforms open-source LLaMA-13B model.

* Corresponding author
E-mail address: jinfuliu@hit.edu.cn (Jinfu Liu)

1. **Introduction**

Nowadays, the overuse of fossil fuels has caused serious environmental problems, such as global warming and sea level rising[1-3]. Developing renewable energy, such as wind energy and solar energy, is one of the most effective methods to solve the environmental problems[4,5]. With the proposal of carbon neutrality goal, the development of renewable energy is accelerating. Developing renewable energy is currently one of the most important for solving environmental problems and achieving carbon neutrality goal, and thus renewable energy is becoming one of the most important academic research fields [6].

In the era of Artificial Intelligence (AI), many researchers have introduced AI algorithms into renewable energy field [7,8]. Many AI techniques, such as deep neural networks, support vector machines, gradient boosting decision trees etc., have been successfully used to forecast the future wind power and photovoltaic power generation [9]. AI techniques like deep neural networks., have also been successfully used in the Prognostics and Health Management (PHM) of renewable energy systems [10]. AI techniques like deep reinforcement learning, have made great success in operational optimization of renewable energy systems [11]. With the large-scale renewable energy in the electricity grid, AI has also been successfully used in the trade of electricity markets[12]. Therefore, AI has been playing an increasingly important role in the development of renewable energy.

As the recent breakthrough and the state-of-the-art technique in AI field, ChatGPT (Chat Generative Pre-trained Transformer) [13] is a Large Language Model (LLM) trained on billions of text data from websites [14,15]. ChatGPT is capable of generating human-like text, answer questions like human, and complete various language-related tasks with high accuracy [16]. Since its launch by OpenAI company in November 2022, it has quickly become one of the fastest-growing consumer applications in history, with millions of active users monthly. Its vast knowledge base and language processing capabilities have revolutionized the way people interact with technology [17], making it easier and more natural to communicate with machines [18]. ChatGPT has been successfully used in various fields, including business, education, manufacturing, healthcare, medicine, and communication. Through ChatGPT, renewable energy researchers can automatically generate some academic paper paragraphs, and improve the academic writing efficiency.

The boom of ChatGPT brings the popularity of Large Language Models (LLMs). To compete with ChatGPT developed by OpenAI, many companies like Google company, Baidu company, Anthropic company, iFLYTEK company etc., are publishing their own LLMs. These new LLMs include Claude model [19] developed by Anthropic company, ERNIE Bot model [20-22] developed by Baidu company, SparkDesk model [23] developed by iFLYTEK company etc. These LLMs can be comparative to ChatGPT in some aspects. Through these LLMs, renewable energy researchers can also generate some useful academic paper paragraphs, and combines the text generated by ChatGPT to further improve the academic writing efficiency.

Although above LLMs have various merits, they are non-open source models. Users can only use them through their official websites. Additionally, above LLMs are not specially designed to satisfy the need for a certain field. When above general LLMs are applied to a certain field, they cannot give the optimal replies due to the lack of accurate and domain-specific expert knowledge [24]. One possible solution to this issue is finetuning the open-source LLMs for a specific field to introduce the domain knowledge to LLMs. LLaMA and GLM (General Language Model) are two popular open-source general LLMs. Touvron et al. [25] proposed open-source LLaMA model, and demonstrated the superiorities over conventional open-source LLMs. Du et al. [26] proposed open-source GLM models and demonstrated good performance. Researchers have finetuned open-source general LLMs for various specific fields. In the field of medicine, there are DoctorGLM [27] open-source model, HuatuoGPT [28] open-source model, Bian Que [29] open-source model etc. In the field of law, there are LawGPT[30] open-source model and Lawyer LLaMA[31] open-source model. In the field of finance, Cornucopia-LLaMA-Fin-Chinese [32] is a good attempt to finetune LLaMA.

All above LLMs for specific domains have achieved better performance than general LLMs in the target domain. However, in the field of renewable energy and carbon neutrality, there has not been any open-source LLMs for this specific domain. Additionally, there has not been any open-source datasets related to renewable energy for training LLMs. The lacking of open-source datasets and domain-specific LLMs greatly restricts the applications of LLMs in renewable energy field. These are the problems this paper deals with.

To address these problems, this paper published the first open-source Renewable Energy Academic Paper (REAP) dataset for non-commercial LLM research of renewable energy, which contains the title and abstract of 1,168, 970 academic papers in Web of Science. Based on REAP dataset, HouYi ("后羿" in Chinese) model, the first LLM for renewable energy, is developed through finetuning general LLM models. The proposed HouYi model demonstrated better performance than general LLMs in renewable energy field.

**The main contributions of this paper are summarized as follows:**

Firstly, this paper published the first open-source dataset called Renewable Energy Academic Paper (REAP) dataset with the title and abstract of 1,168, 970 academic papers for training LLMs on renewable energy field. To the best of our knowledge, this is the first time that the open-source dataset for training LLMs in renewable energy field has been proposed.

Secondly, HouYi ("后羿" in Chinese) model, the first LLM for renewable energy field is developed through finetuning general LLMs. A public web demo for HouYi model of renewable energy field is published, which is available to all readers. To the best of our knowledge, this is the first open-source LLM that specially designed for renewable energy fields.

Thirdly, this paper introduces human experts-based analytic hierarchy process (AHP) method evaluate LLMs. To the best of our knowledge, this is the first time that AHP method has been used for evaluating LLMs. Experiments verify that HouYi model has excellent abilities in paper generation of renewable energy field.

The rest of this paper is organized as follows. Section 2 constructs the REAP dataset for renewable energy. Section 3 gives the methods and procedure for building HouYi model. Section 4 performs academic paper generation experiments and evaluates its performance. Section 5 concludes the paper and outlines the future research orientation.

## 2. Construction of Renewable Energy Academic Paper (REAP) dataset

Renewable energy includes wind energy, solar energy, Hydrogen energy, geothermal energy, biomass energy etc. Compared with conventional fossil fuels, the power generation using renewable energy can fluctuate with time. This makes the demands for energy storage increase. Through the integration of renewable energy and energy storage, it is promising to realize carbon neutrality goal. Therefore, this paper used a series of related keywords including "sustainable energy", "wind energy", "solar energy", "photovoltaic power", "Hydrogen energy", "biomass energy", "geothermal energy", "energy storage", "photothermal", "carbon neutrality", "Energy Internet", "virtual power plant" etc., to search literatures in Web of Science database.

In the experiments, 1,168, 970 literatures were obtained. This paper used the title of these literatures as the input and the abstract of these literatures as the output to train LLMs, and thus proposed the first open-source Renewable Energy Academic Paper (REAP) dataset for non-commercial LLM research of renewable energy. An example of the input and output is given in Table 1, which is extracted from literature [33].

Table 1 An example of input and output of REAP dataset

| Input | Hybrid forecasting model-based data mining and genetic algorithm-adaptive particle swarm optimisation: a case study of wind speed time series |
|---|---|
| Output | Wind energy has been part of the fastest growing renewable energy sources and is clean and pollution-free. Wind energy has been gaining increasing global attention, and wind speed forecasting plays a vital role in the wind energy field. However, such forecasting has been demonstrated to be a challenging task due to the effect of various meteorological factors. This study proposes a hybrid forecasting model that can effectively provide preprocessing for the original data and improve forecasting accuracy. The developed model applies a genetic algorithm-adaptive particle swarm optimisation algorithm to optimise the parameters of the wavelet neural network (WNN) model. The proposed hybrid method is subsequently examined in regard to the wind farms of eastern China. The forecasting performance demonstrates that the developed model is better than some traditional models (for example, back propagation, WNN, fuzzy neural network, and support vector machine), and its applicability is further verified by the paired-sample T tests. |

REAP dataset contains 1,168, 970 academic literatures. Among these literatures, there are 711, 126 journal papers, 347, 616 patents, 109, 419 conference papers, 18 books and 791 books in series. The literature type information of these literatures is given in Fig. 1. For all the 711, 126 journal papers, we list the top 20 journal names and the number of articles for each journal in Fig. 2.

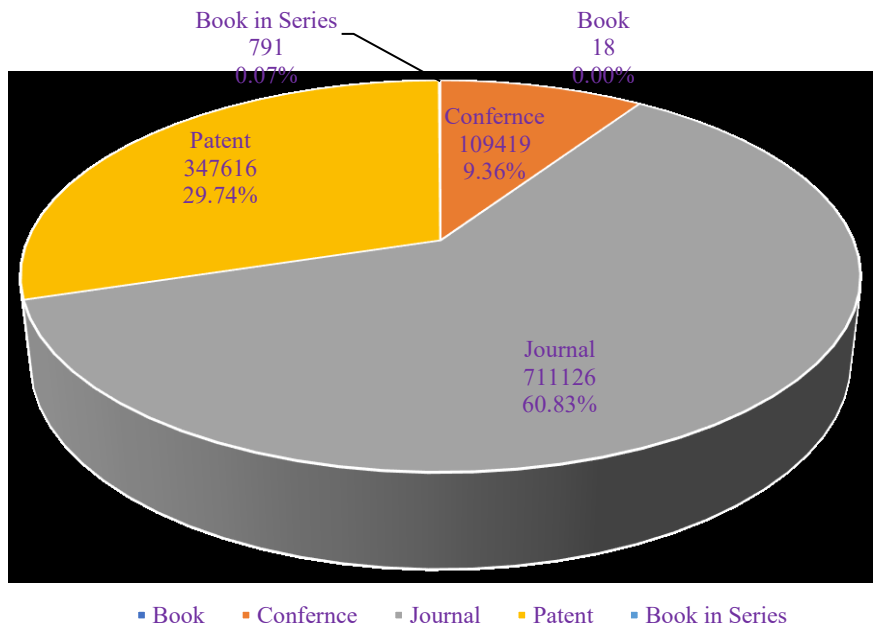

Fig. 1 Literature type distribution of REAP dataset

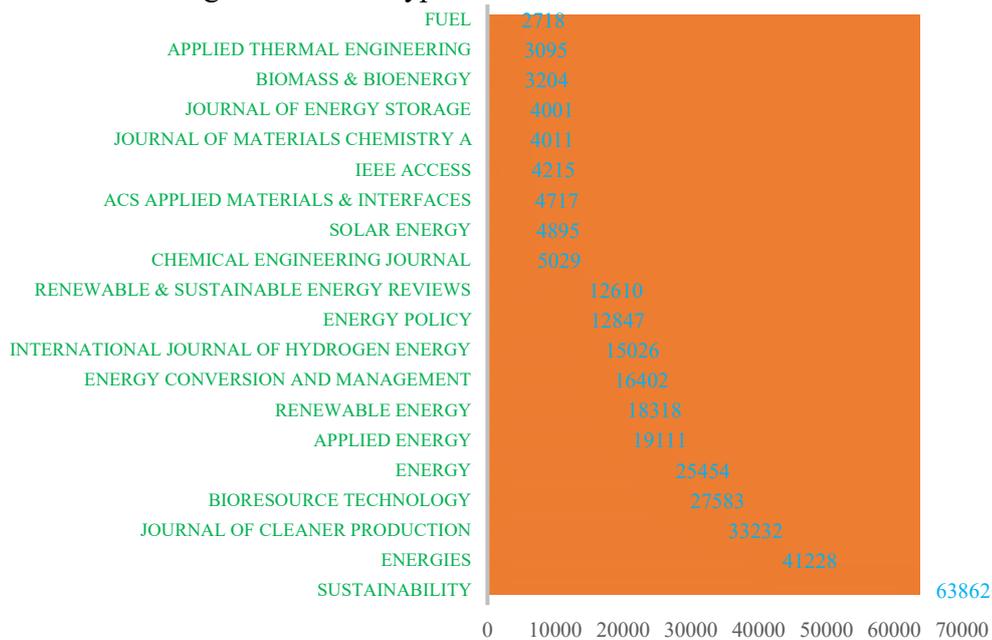

Fig. 2 Top 20 journals in REAP dataset

It is observed in Fig. 1 that REAP dataset contains many journal papers, conference papers and patents in renewable energy field. From Fig. 2, it is observed that papers from many famous energy journals including Energy Conversion and Management, Energy, Solar Energy, Renewable & Sustainable Energy Reviews, Journal of Energy Storage, Applied Thermal Engineering, Renewable Energy, Applied Energy etc., are included in the REAP dataset. Therefore, it is reasonable to believe that REAP dataset can provide professional information in renewable energy field. The REAP dataset is available at https://github.com/mingliangbai/HouYi and will be public immediately after acceptance.

## 3. Methods
### 3.1. Large language model

Large language models (LLMs) are trained using a huge amount of text data including website data, social media data and document data. LLMs can generate human-like responses to prompts. Given a question or command, LLMs can answer it like humans. ChatGPT is currently the most popular LLM. Fig. 3 shows an example of ChatGPT's responses.

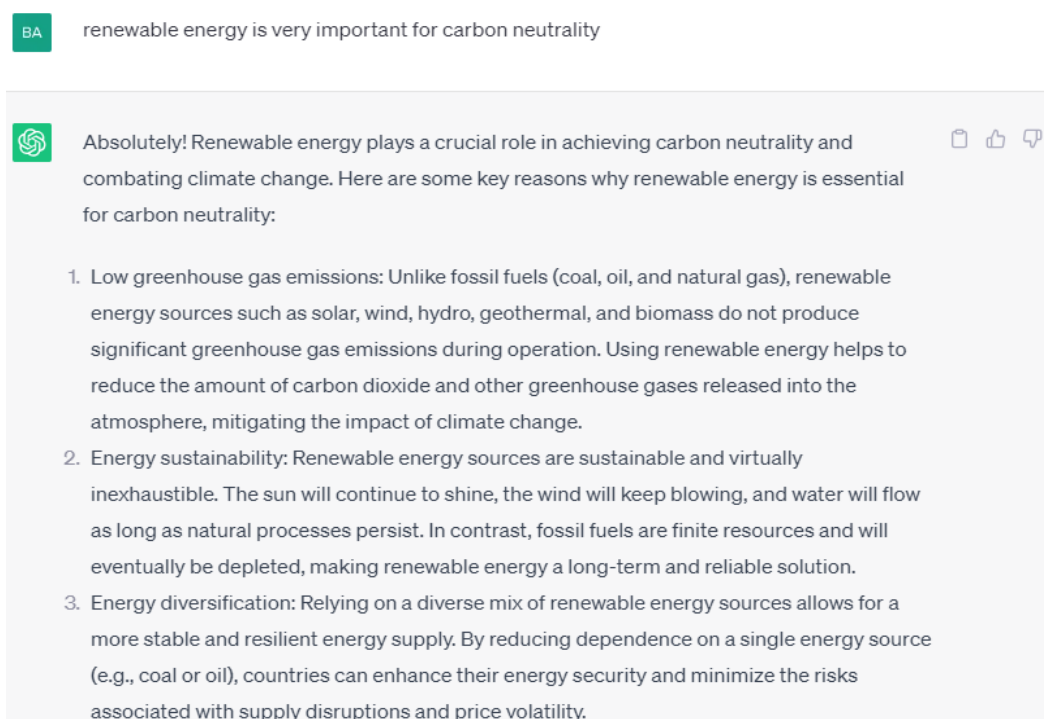

Fig. 3 An example of responses generated by ChatGPT

It is observed from Fig. 3 that ChatGPT can help renewable energy researchers with academic writing. However, ChatGPT is close-source and is a general LLM without paying sufficient attention to renewable energy field. Thus, this paper will train an open-source LLM specially designed for renewable energy field.

### 3.2. Transformer

The basic network component in LLMs is the Transformer neural network.

Transformer is first proposed by Google company in literature [34]. Compared with conventional recurrent neural networks or convolutional neural networks, Transformer is more parallelizable and requiring significantly less time to train. Currently, Transformer has become one of the most popular methods in natural language processing. Transformer is composed of several encoders and decoders. For each encoder, there is a corresponding decoder. The principle of Transformer is shown in Fig. 4. The detailed structure in each encoder and decoder of Transformer is shown in Fig. 5.

The encoder is composed of a stack of $N$ identical layers. Each layer has two sub-layers. The first is a multi-head self-attention mechanism, and the second is a simple, position-wise fully connected feed-forward network. Residual connection and layer normalization are used in "add & norm" part shown in Fig. 5. The principle of attention and feed-forward network are given in Equation (1) and Equation (2) respectively.

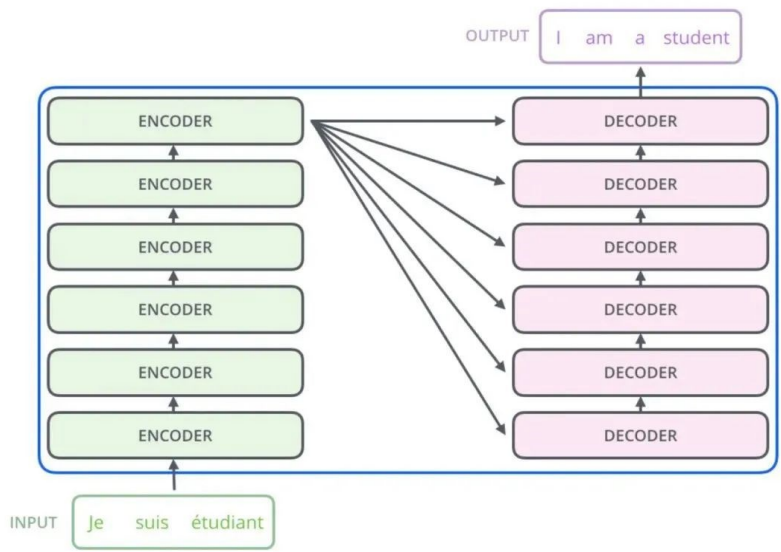

Fig. 4 Encoder and decoder structure of Transformer

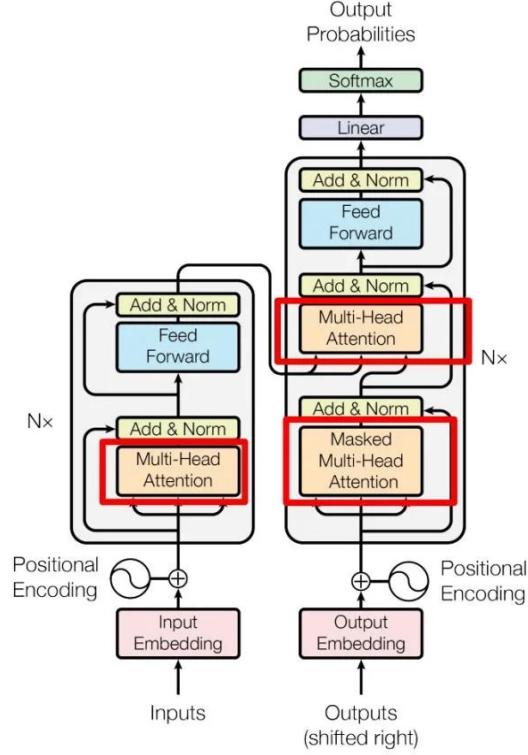

Fig. 5 Detailed principles of Transformer

$$\text{Attention}(Q, K, V) = \text{softmax}\left(\frac{QK^T}{\sqrt{d_k}}\right)V \quad (1)$$

$$\text{FFN}(x) = \max(0, xW_1 + b_1)W_2 + b_2 \quad (2)$$

where $Q$ is the query, $K$ is the key, $V$ is the value and $d_k$ is the dimension of keys. $W_1$ and $W_2$ are the weight parameters, $b_1$ and $b_2$ are the bias parameters, and $x$ is the input of this layer.

The decoder is also composed of a stack of $N$ identical layers. In addition to the two sub-layers in each encoder layer, the decoder inserts a third sub-layer, which performs multi-head attention over the output of the encoder stack. Through the encoder-decoder architecture and multi-head attention mechanism, Transformer can successfully extract the context information, and thus has become the most popular methods in natural language processing and the most important network component in LLMs.

**3.3 General Language Model (GLM)**

General Language Model (GLM) [35] is a popular open-source LLM that can address natural language understanding, unconditional generation and conditional generation problem well simultaneously. GLM takes advantages of autoencoding models, autoregressive models and encoder-decoder models, and has shown great advantages in various natural language processing tasks. The principle of GLM is shown in Fig. 6.

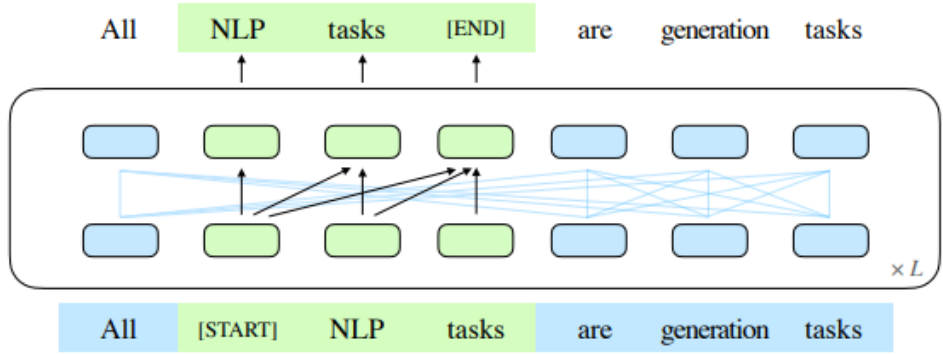

Fig. 6 Principle of GLM [35]

The training goal of GLM is to fill the missing tokens in the spans from the corrupted text. Given an input text $[x_1, \ldots, x_n]$, multiple text spans $\{s_1, \ldots, s_m\}$ are sampled, where each span $s_i$ corresponds to a series of consecutive tokens $[s_{i,1}, \ldots, s_{i,l_i}]$ in $x$. Let $Z_m$ be the set of all possible permutations of the length-$m$ index sequence $[1, 2, \ldots, m]$, and $s_{z_{<i}}$ be $[s_{z_1}, \ldots, s_{z_{i-1}}]$, the pretraining objective is as follows:

$$\max_{\theta} \mathbb{E}\left[\log p_\theta\left(s_{z_i} \mid x_{\text{corrupt}}, s_{z_{<i}}\right)\right] \qquad (3)$$

Through multi-task pretraining, GLM can perform well in various tasks. Based on the compression of GLM, an open-source model with 6 billion parameters called ChatGLM-6B model[36] has been established and become a commonly-used open-source LLM.

**3.4. Finetune transfer learning**

GLM is a general LLM without paying enough attention on a specific domain. Thus, GLM cannot guarantee the optimal performance when applying to the renewable energy field. This paper uses the renewable energy domain data, namely the REAP dataset proposed in Section 2 to finetune ChatGLM-6B model and establish an LLM specifically designed for renewable energy and carbon neutrality field.

Finetune transfer learning is a popular method in deep learning. Given a source domain $D_s$ with $n$ labelled data $D_s = \{x_s, y_s\}_{s=1}^{n}$ and a target domain $D_t$ with $m$ labelled data $D_t = \{x_t, y_t\}_{t=1}^{n}$, where the marginal probability distribution are different, namely $P_s(x_s) \neq P_t(x_t)$ and the conditional probability distribution are also different, namely $Q_s(y_s/x_s) \neq Q_t(y_t/x_t)$, then the goal of finetune transfer learning is utilizing the data from $D_s$ to help improve the performance of $D_t$, namely obtaining the neural network parameter $\theta$ satisfying the following equation:

$$\theta = \arg\max P(y_t | x_t) \tag{4}$$

In deep neural networks like Transformers, the features are extracted layer-by-layer. Fig. 7 illustrates the layer-wise feature extraction process. The shadow layers or the first several layers can extract general features like edges or the combination of edges. The deep layers or the last several layers can extract specific features, like object models. Thus, the common way for finetune transfer learning is to froze the first serval layers and finetune the rest layers to avoid training from scratch and improve the network performance.

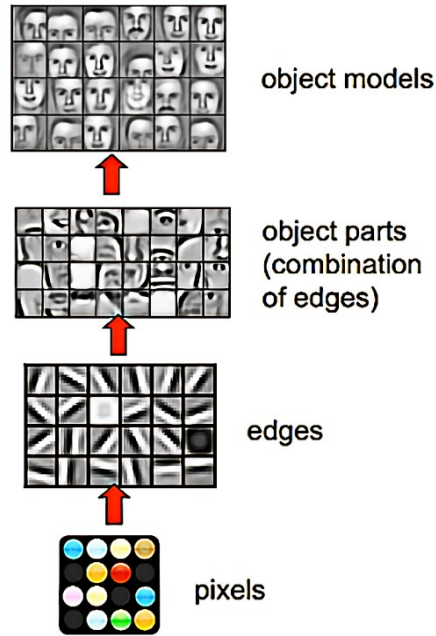

Fig. 7 layer-wise feature selection

This paper finetuned ChatGLM-6B model using the proposed REAP dataset, whose inputs and outputs are the title and abstract of academic papers in renewable energy field. During the finetuning process, the network is trained on a computer running Ubuntu 20.04 system with eight RTX 3090 graphic cards, the learning rate is 0.01 and the leaning epoch is 30. Thus, HouYi ("后羿" in Chinese) model, the first LLM for renewable energy field is developed through finetuning general LLMs. Gradio library in Python is used to build a web demo shown in Fig. 8 for HouYi model.

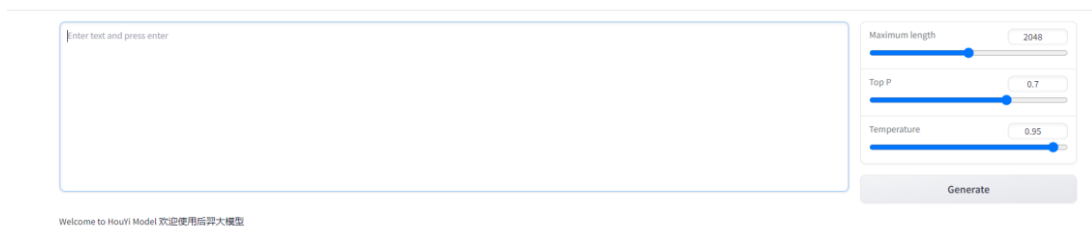

Fig. 8 Web demo of HouYi model

## 4. Experiments
### 4.1 Paper generation experiments

This paper used the developed HouYi ("后羿" in Chinese) model for generating articles in renewable energy and carbon neutrality field. Two examples of wind energy field and solar energy field are used for evaluating the performance. The proposed

HouYi model is compared with four popular close-source LLMs including ChatGPT[13] developed by OpenAI, Claude [19] model developed by Anthropic company, ERNIE Bot [22] model developed by Baidu company, SparkDesk [23] model developed by iFLYTEK company[23]. Meanwhile, the proposed HouYi model is also compared with open-source LLaMa-13B model developed by Meta company. The paper generation results of above six LLMs for wind energy field and solar energy field are given in Table 2 and Table 3 respectively.

Table 2 Paper generation experiment for wind energy field

| LLM model | Input: write a paper entitled "Uncertainty estimation for wind energy conversion by probabilistic wind turbine power curve modelling" |
|---|---|
| **HouYi** | Title: Uncertainty Estimation for Wind Energy conversion by probabilistic Wind turbine Power Curve Modelling<br>Abstract:<br>    In this paper, we provide an overview of the uncertainty estimation for wind energy conversion by probabilistic wind turbine power curve modelling. We discuss the background of the model, the method of uncertainty analysis, and the results of the model. Additionally, we present a case study of a wind power plant with a probabilistic wind turbine power curve modelling to illustrate the method.<br>Introduction:<br>    Wind energy is one of the most promising sources of renewable energy, which has been increasing in recent years. However, the conversion of wind energy into electricity is not always straightforward, as it involves a complex system of equipment and processes. The uncertainty of the conversion process can be significant, and it is essential to estimate the uncertainty of the conversion process to ensure that it is within the allowed range.<br>    Probabilistic wind turbine power curve modelling is a useful tool for uncertainty estimation. It models the output power of a wind turbine based on the input power, the wind speed, and the weather conditions. The model can provide an uncertainty analysis of the output power and it can also be used to optimize the output power of the wind turbine based on the input power and weather conditions.<br>Methodology:<br>    To estimate the uncertainty of the wind energy conversion process, we use a probabilistic wind turbine power curve modelling method. We model the output power of a wind turbine based on the input power, the wind speed, and the weather conditions. The model is based on the power law, which states that the output power of a wind turbine is directly proportional to the input power. The model is based on the following parameters: input power, wind speed, weather conditions<br>    The input power is used to determine the output power of the wind turbine. The output power is then used to determine the power factor. The power factor is a measure of the efficiency of the wind turbine. The power factor is determined by the load and the wind speed. The power factor is used to determine the output power of the wind turbine.<br>    The model is based on the following steps: |
| | 1. Estimation of the input power: The input power is determined based on the weather conditions and the load. The input power is used to determine the output power of the wind turbine.

2. Estimation of the output power: The output power is determined based on the input power, the wind speed, and the weather conditions. The output power is used to determine the power factor.

3. Estimation of the power factor: The power factor is determined based on the input power, the wind speed, and the weather conditions. The power factor is used to determine the output power of the wind turbine.

4. Estimation of the uncertainty of the output power: The uncertainty of the output power is determined based on the input power, the wind speed, and the weather conditions. The uncertainty is used to determine the uncertainty of the output power of the wind turbine.

5. Estimation of the uncertainty of the power factor: The uncertainty of the power factor is determined based on the input power, the wind speed, and the weather conditions. The uncertainty is used to determine the uncertainty of the power factor of the wind turbine.

Case Study:

To illustrate the method, we present a case study of a wind power plant with a probabilistic wind turbine power curve modelling. The plant has a capacity of 2000MW, and it is located in the Suburb of a city in China. The plant is equipped with 30 wind turbines, each of which has a capacity of 100MW. The wind speed is not constant in the suburb, it can vary from 30 to 50 m/s. The conditions are not constant, it can vary from 10 to 40°C in the summer and from 10 to 30°C in the winter.

The company decided to convert the wind energy into electricity and they used the probabilistic wind turbine power curve modelling to optimize the output power of the wind turbine based on the input power and weather conditions. The company used the following model:

output power = 0.8 × input power + 0.1 × wind speed + 0.1 × power factor

The model was run 100 times and the output power was measured for each run. The results are shown in the table below:

| % uncertainty | % uncertainty | % uncertainty | % uncertainty |
|---|---|---|---|
| 0.5 | 0.5 | 0.5 | 0.5 |
| 1.0 | 1.0 | 1.0 | 1.0 |
| 2.0 | 2.0 | 2.0 | 2.0 |

The uncertainty of the output power is 0.5% for each run and it is also 0.5% for each run. The uncertainty of the power factor is also 0.5% for each run and it is also 0.5% for each run. The uncertainty of the output power is 1% for the first run and it is also 1% for the second run. The uncertainty of the power factor is 1% for the first run and it is also 1% for the second run.

Conclusion:

In this paper, we provide an overview of the uncertainty estimation for wind energy conversion by probabilistic wind turbine power curve modelling. We discuss the background of the model, the method of uncertainty analysis, and the results of the model. Additionally, we present a case study of a wind power plant with a probabilistic wind turbine power curve modelling to illustrate the method. The method can provide an uncertainty analysis of the output power of the wind turbine and it can also be used to optimize the output power of the wind turbine based on the input power and weather conditions. The uncertainty of the output power is 0.5% for each run and it is also 0.5% for each run. The uncertainty of the power factor is also 0.5% for each run and it is also 0.5% for each run. The method can be used in many other countries to estimate the uncertainty of the conversion process. |
| Claude | Here is a possible outline for a paper on "Uncertainty Estimation for Wind Energy Conversion by Probabilistic Wind |
| | Turbine Power Curve Modelling":

1. Introduction

- Thesis: Probabilistic power curve models quantify the uncertainty in wind turbine power generation by estimating the distribution of power output at different wind speeds, providing more realistic estimates of wind energy potential.

2. Deterministic vs Probabilistic Wind Turbine Power Curve Models

- Review conventional (deterministic) power curve models: assumed 1-to-1 mapping between wind speed and power

- Limitations of deterministic models in capturing variability and uncertainty in power generation

- Overview of probabilistic power curve models: estimate distributions of power output at different wind speeds to reflect uncertainty

3. Sources of Uncertainty in Wind Turbine Power Generation

- Variability in wind speed during power production

- Turbulence intensity changing mechanical loads and power conversion

- Modelling uncertainties in aerodynamic models

- Manufacturing tolerances and component reliability issues

- Control/operational uncertainties due to wind farm controls and grid demands

4. Modelling Methodologies for Probabilistic Power Curves

- Basic statistics: mean, quantiles/percentiles, standard deviation, distribution fitting

- Kernel density estimation

- Monte Carlo simulation

- Machine learning: Gaussian processes, neural networks, etc.

- Discuss strengths and limitations of different methods

5. Case Studies

- Apply different probabilistic power curve modelling techniques on sample wind turbine data

- Evaluate their computational efficiency and accuracy in capturing power output distributions

- Discuss their ability to link power variability to different underlying sources of uncertainty

6. Applications and Future Work

- Use in more realistic energy yield assessment and project feasibility studies

- Integration with system models for operational forecasting and optimization

- Fusing with ensemble weather forecasts to provide probabilistic power predictions

- Importance of understanding joint distributions between multiple wind turbines

- Opportunities to develop standardized methods for quantifying and communicating uncertainty

7. Summary and Conclusions

References

    This outline covers the main elements I would suggest including in a paper on this topic. The details under each section offer possibilities to explore. Please adapt the content and structure as needed to best present your key messages. The paper could analyze suitable techniques for probabilistic power curve modelling, evaluate them using case studies, discuss ways to link different sources of uncertainty to power output variability, and explore potential applications. Discussing limitations of existing standards and value of fusing multiple methods could provide useful insights. Please let me know if you have any questions or want any help developing this concept further! I would be happy to assist you. |
| ChatGPT | Title: Uncertainty Estimation for Wind Energy Conversion by Probabilistic Wind Turbine Power Curve Modelling

Abstract:

    With the growing emphasis on renewable energy sources, wind energy conversion systems, particularly wind |

| | |
|---|---|
| | turbines, have gained significant attention in recent years. However, accurately predicting the power output of wind turbines remains challenging due to the inherent uncertainty associated with wind speed measurements and turbine performance characteristics. This paper proposes a novel approach to estimate the uncertainty in wind energy conversion by employing probabilistic modelling techniques to construct wind turbine power curves. By incorporating uncertainty estimation into the power curve modelling process, more reliable predictions of wind turbine performance can be obtained, facilitating better decision-making in wind energy projects.<br><br>Introduction<br>1.1 Background<br>1.2 Motivation<br>1.3 Objectives<br>Uncertainty in Wind Energy Conversion<br>2.1 Sources of Uncertainty<br>2.2 Impact of Uncertainty on Power Output<br>2.3 Need for Uncertainty Estimation<br>Probabilistic Modelling of Wind Turbine Power Curves<br>3.1 Traditional Power Curve Modelling<br>3.2 Probabilistic Power Curve Modelling<br>3.3 Benefits of Probabilistic Modelling<br>Methodology<br>4.1 Data Collection and Preprocessing<br>4.2 Construction of Probabilistic Power Curves<br>4.3 Uncertainty Estimation Techniques<br>4.4 Validation and Calibration<br>Case Study: Uncertainty Estimation in Wind Energy Conversion<br>5.1 Data Description<br>5.2 Probabilistic Modelling Results<br>5.3 Comparison with Traditional Power Curve Models<br>5.4 Analysis of Uncertainty Effects<br>Implications and Applications<br>6.1 Impact on Wind Farm Performance Assessment<br>6.2 Optimization of Wind Turbine Operations<br>6.3 Decision-Making in Wind Energy Projects<br>Challenges and Future Directions<br>7.1 Data Availability and Quality<br>7.2 Advanced Uncertainty Estimation Techniques<br>7.3 Integration with Other Wind Energy Models<br>Conclusion<br>References<br>Keywords: Wind energy conversion, wind turbine, uncertainty estimation, probabilistic modelling, power curve, renewable energy. |
| ERNIE Bot | Uncertainty estimation for wind energy conversion by probabilistic wind turbine power curve modelling<br><br>  With the increasing importance of renewable energy in the energy mix, wind energy conversion technology has |
| | become a focus of academic and industrial research. Wind turbine power curves are important tools for accurately predicting wind power generation, describing the relationship between wind turbine output power and wind speed. However, due to the randomness and uncertainty of wind speed, there are certain difficulties in modelling wind turbine power curves. This article aims to explore the probabilistic wind turbine power curve modelling technique based on probability methods and evaluate the uncertainty during modelling.

In recent years, wind turbine power curve modelling methods have mainly included deterministic methods and probabilistic methods. Deterministic methods model the functional relationship between wind power generation and wind speed by fitting experimental data, such as polynomial functions and exponential functions. However, deterministic methods do not consider the randomness and uncertainty of wind speed and therefore have certain limitations in practical applications. Probabilistic methods take into account the randomness of wind speed and model the relationship between wind speed and wind power generation probability distribution, such as Weibull distribution and Gamma distribution.

In this article, a probabilistic wind turbine power curve model is established, with the specific steps as follows:

Collect actual output power data of wind turbines under different wind speeds and analyze and process them.

Select an appropriate probability distribution to describe the relationship between wind speed and wind power generation based on the statistical characteristics of wind speed and wind power generation.

Use maximum likelihood estimation and other methods to estimate the parameters of the probability distribution and establish the wind turbine power curve model.

Perform performance evaluation of the wind turbine power curve model, such as root mean square error and mean absolute error.

In this article, actual wind power generation data were collected, and a Weibull distribution was used to describe the relationship between wind speed and wind power generation. Through maximum likelihood estimation, the parameters of the Weibull distribution were obtained, and the wind turbine power curve model was established. The performance evaluation results show that the model established in this article has good prediction accuracy and stability.

The probabilistic wind turbine power curve model established in this article can effectively describe the relationship between wind speed and wind power generation and has high prediction accuracy and stability. At the same time, this article evaluates the uncertainty during modelling, providing a reference for further development and optimization of wind energy conversion technology. Future research can further explore the applicability of different probability distributions and how to reduce the impact of modelling uncertainty to improve wind energy conversion efficiency. |
| SparkDesk | Title: Uncertainty Estimation for Wind Energy Conversion by Probabilistic Wind Turbine Power Curve Modelling

Abstract:

Wind energy is a promising source of renewable energy with great potential to address the global energy demand. However, wind turbines are highly sensitive to various factors such as wind speed, turbulence, and load, which can lead to variations in power output. This paper presents a probabilistic wind turbine power curve modelling approach for estimating uncertainty in wind energy conversion. The proposed method involves building a probabilistic model that captures the complex relationship between wind turbine parameters and power output. By integrating data from multiple sensors and wind farms, the model can provide accurate predictions of wind power production under different conditions.

The paper also discusses the challenges associated with this approach, including the need for high-quality data and the importance of incorporating external factors such as environmental conditions and weather forecasts. Finally, the paper highlights the potential benefits of using probabilistic modelling for wind energy conversion, including improved reliability and efficiency of wind power systems.

1. Introduction

1.1 Background on wind energy |

| | |
|---|---|
| | 1.2 Challenges in wind energy conversion |
| | 1.3 Importance of uncertainty estimation |
| | 2. Probabilistic Wind Turbine Power Curve Modelling |
| | 2.1 Overview of the modelling approach |
| | 2.2 Building the probability distribution function (PDF) |
| | 2.3 Incorporating external factors into the model |
| | 2.4 Using historical data to train the model |
| | 2.5 Predicting wind power production under different conditions |
| | 3. Case Study: Application of Probabilistic Wind Turbine Power Curve Modelling in a Wind Farm |
| | 3.1 Data collection from a wind farm |
| | 3.2 Building the probabilistic model for the wind farm |
| | 3.3 Evaluating the performance of the model using metrics such as mean absolute error (MAE) and root mean square error (RMSE) |
| | 3.4 Discussion of the results and their implications for wind energy conversion |
| | 4. Challenges and Future Work |
| | 4.1 Challenges associated with the modelling approach |
| | 4.2 Possible improvements and extensions to the model |
| | 4.3 Potential applications of probabilistic modelling in other areas of wind energy research and development |
| | 5. Conclusion |
| | 5.1 Summary of the main findings |
| | 5.2 Implications for wind energy conversion and sustainability efforts |
| LLaMA-13B | Firstly, let's look at the abstract, which is a summary of your paper.<br><br>The abstract should be no more than 300 words, and should give the reader a clear overview of the paper's objectives, main results and conclusions. Ideally, an abstract should be able to stand alone as an accurate description of your paper. The abstract should be written at the end of the paper, and should contain the following information:<br><br>● The main objectives of your work.<br>● A summary of your methodology.<br>● The main results of your work.<br>● The main conclusions of your work.<br><br>Now let's look at the introduction.<br><br>The introduction should give an overview of your research. It should describe the main objective of your research, its context, and its motivation.<br><br>You should also describe the problem you set out to solve and the scientific principles that are relevant to your work. The introduction should also explain the scientific and technical background to your work, including any previous work that you have done in this area.<br><br>It is important to give a complete overview of the field, so the reader can understand the significance of your research. Your introduction should be no more than 300 words.<br><br>The introduction should be written after the abstract, and should contain the following information:<br><br>● Description of the main objective of your research.<br>● Description of the problem you set out to solve.<br>● Description of the scientific principles that are relevant to your research.<br>● Description of the technical background to your work, including any previous work that you have done in this |
|  | area.<br><br>Now let's look at the main body of the paper. The main body of the paper should contain all the details of your research.<br><br>The main body of the paper should be no more than 2000 words.<br><br>The main body of the paper should contain the following information:<br><br>● Description of the methodology used, including the main assumptions made, and how the results were obtained.<br>● Description of the main results of your research, with relevant figures and tables.<br>● Description of any conclusions that can be made from your research.<br>● Description of any limitations of your research, indicating which areas would benefit from further research.<br><br>Now let's look at the conclusion. The conclusion should give a summary of your research and the main results. The conclusion. |

Table 3 Paper generation experiment for solar energy

|  |  |
|---|---|
|  | write a paper entitled "A comparative study on machine learning methods-based numerical weather prediction (NWP) post-processing for one-day-ahead photovoltaic power forecasting" |
| HouYi | Title: A comparative study on machine learning methods-based numerical weather prediction (NWP) post-processing for one-day-ahead photovoltaic power forecasting<br><br>Abstract:<br><br>This paper presents a comparative study on the performance of machine learning methods based numerical weather prediction (NWP) post-processing for one-day-ahead photovoltaic power forecasting. The study is conducted using a real-world case study in which a PV module is connected to a power grid. The goal of the study is to compare the effectiveness of different machine learning methods for improving the accuracy of the PV power forecasting model. The results of the study show that the use of deep learning methods, such as convolutional neural networks (CNNs), can provide significant improvements in the accuracy of the power forecasting model. The use of other machine learning methods, such as support vector machines (SVMs), does not provide significant improvements in the accuracy of the model. The study also demonstrates that the use of a combination of both CNNs and SVMs can provide the best performance for the power forecasting model.<br><br>Introduction:<br><br>PV power forecasting is a critical aspect of renewable energy production and management. The accuracy of the forecast is crucial for ensuring the safe and reliable operation of PV power systems. Nowadays, many machine learning methods have been developed to improve the accuracy of PV power forecasting models. These methods are based on the use of advanced sensors and artificial intelligence (AI) techniques, which can provide more accurate and faster results than traditional methods.<br><br>numerical weather prediction (NWP) is a type of weather forecasting that uses computer models to predict the weather patterns of different regions. WP models are based on the National Oceanic and Atmospheric Administration (NOAA) Advanced Neural Networks (ANN) model. The use of ANNs can provide more accurate and faster results than traditional methods, especially for complex weather patterns.<br><br>In this study, we compare the effectiveness of different machine learning methods for improving the accuracy of one-day-ahead photovoltaic power forecasting models. The study is conducted using a real-world case study in which a PV module is connected to a power grid. The goal of the study is to compare the effectiveness of different machine learning methods for improving the accuracy of the power forecasting model.<br><br>Methodology:<br><br>To conduct this study, we use the same PV power forecasting model that was used in the real-world case study. |
| | The model is based on the Advanced Neural Networks (ANN) model developed by NOAA. The model is trained using historical data of PV power production in a specific region. The training data includes the weather data, PV module data, and power production data.

We compare the effectiveness of different machine learning methods for improving the accuracy of the power forecasting model. The methods include:

Deep Learning methods: CNNs are a type of deep learning method that can provide significant improvements in the accuracy of the power forecasting model.

SVMs: SVMs are a type of machine learning method that can provide significant improvements in the accuracy of the power forecasting model.

Traditional methods: The traditional methods include the use of regression models, such as linear regression, and decision trees.

We use a leave-one-out cross-validation (LOOCV) method to evaluate the performance of the power forecasting model. The LOOCV method involves training the model on a training data set, testing the model on a test data set, and evaluating the performance on the test data set. We compare the performance of the power forecasting model using the LOOCV method with the performance of the same model using traditional methods.

Results:

The results of the study show that the use of deep learning methods, such as CNNs, can provide significant improvements in the accuracy of the power forecasting model. The use of CNNs and SVMs can also provide good performance for the power forecasting model. However, the use of both CNNs and SVMs can provide the best performance for the power forecasting model.

Conclusion:

In conclusion, the use of deep learning methods, such as CNNs, can provide significant improvements in the accuracy of one-day-ahead photovoltaic power forecasting models. The use of SVMs can also provide good performance for the power forecasting model. The use of both CNNs and SVMs can provide the best performance for the power forecasting model. The study also demonstrates that the use of a combination of both CNNs and SVMs can provide the best performance for the power forecasting model.

Future work:

Future work can involve developing more advanced CNNs and SVMs for PV power forecasting, and testing them on real-world data sets to evaluate their performance. The use of advanced machine learning methods can also be explored for other renewable energy systems, such as wind and solar power forecasting. |
| Claude | Here is a possible outline for the paper:

1. Introduction

- Thesis: Machine learning methods can be effective techniques for post-processing numerical weather prediction models and improving short-term PV power forecasting accuracy.

2. Overview of PV Power Forecasting

- Importance of PV power forecasting for power system operation

- Review of different forecasting approaches: physical, statistical and hybrid models

- Error sources in day-ahead PV power forecasting and role of NWP post-processing

3. Numerical Weather Prediction (NWP) Models

- Working principle of NWP models in simulating weather variables

- Common NWP models: GFS, ECMWF, WRF, etc.

- Pros and cons of different models for solar forecasting applications |

- Sources of error in NWP models: inaccurate initial conditions, coarse resolution, imperfect physics, etc.

4. Survey of Machine Learning Methods

- Linear regression

- Decision trees

- Neural networks

- Support vector regression

- Random forests

- Light gradient boosting machine

- Discuss features and hyperparameters tuning of different ML models

5. Modelling Methodology & Results

- Select sample day-ahead PV power forecasting datasets with NWP models

- Apply multiple ML methods for post-processing the NWP outputs

- Evaluate the ML models' performance in improving forecast accuracy

- Discuss computational efficiency, parameter sensitivity, etc. of different models

6. Discussions

- Compare the results from 5 and discuss the strengths and limitations of each ML method

- Analyze the sensitivity of model performance to location, weather type, forecast horizon, etc.

- Evaluate the value of model ensembles and other advanced techniques

7. Conclusions and Future Work

References

The outline covers the essential elements to discuss in the suggested paper title. Feel free to adapt the structure and content as needed to best present your key messages. The details under each section provide possibilities to explore.

The paper could evaluate and compare suitable ML methods for NWP post-processing in PV power forecasting, discuss ways to improve their performance, and analyze open research questions. Discussing applications of these techniques and their value to the solar industry would also be of interest to readers. Please let me know if you have any questions or need any help further developing this topic! I would be happy to discuss with you.

| | |
|---|---|
| ChatGPT | Title: A Comparative Study on Machine Learning Methods-Based Numerical Weather Prediction (NWP) Post-Processing for One-Day-Ahead Photovoltaic Power Forecasting |

Abstract:

Accurate forecasting of photovoltaic (PV) power generation is crucial for effective grid management and integration of renewable energy sources. Numerical weather prediction (NWP) models are commonly used to provide weather input for PV power forecasting. However, NWP outputs often require post-processing to improve their accuracy. In recent years, machine learning (ML) techniques have shown promising results in enhancing NWP-based PV power forecasts. This paper presents a comparative study that evaluates and compares different ML methods for post-processing NWP outputs to enhance one-day-ahead PV power forecasting.

Introduction:

1.1 Background

1.2 Objectives

1.3 Contribution

Methodology:

2.1 Data Collection and Preprocessing

| | |
|---|---|
| | 2.2 NWP Model Selection and Setup

2.3 Machine Learning Methods

2.3.1 Support Vector Regression (SVR)

2.3.2 Random Forest (RF)

2.3.3 Neural Networks (NN)

2.3.4 Gradient Boosting (GB)

2.4 Feature Selection and Engineering

2.5 Training and Testing Setup

Experimental Results and Analysis:

3.1 Performance Metrics

3.2 Evaluation of Machine Learning Methods

3.3 Impact of Feature Selection and Engineering

3.4 Comparison with Baseline NWP Forecasts

Discussion:

4.1 Interpretation of Results

4.2 Advantages and Limitations of ML Methods

4.3 Practical Implications and Applications

4.4 Future Directions

Conclusion:

5.1 Summary of Findings

5.2 Implications of the Study

5.3 Recommendations for Future Research

References:

(Include a list of relevant sources cited throughout the paper)

Appendix:

(Include any supplementary information or additional details, such as the description of the dataset used, NWP model configurations, and ML model hyperparameters)

Note: This paper outline provides a general structure for your research paper. Make sure to adapt it based on the specific details of your study and the guidelines provided by your institution or intended publication venue. |
| ERNIE Bot | Title: A Comparative Study on Machine Learning Methods-based Numerical Weather Prediction (NWP) Post-processing for One-day-ahead Photovoltaic Power Forecasting

Abstract:

  This paper aims to improve the accuracy of solar power forecasting for one day ahead through post-processing of numerical weather predictions using different machine learning methods. By selecting appropriate machine learning algorithms and adjusting their parameters, we hope to achieve better prediction results. Firstly, we introduce the background and current status of solar power forecasting research. Then, we introduced various machine learning methods' principles and applications. Next, we propose our approach, including data pre-processing, model selection and parameter adjustment, and evaluation of prediction results. Finally, we conduct a small-scale experiment and analyze its results. The experimental results show that different machine learning methods perform differently when dealing with numerical weather predictions, with certain methods significantly improving prediction accuracy under certain conditions.

Introduction: |

|  | With the popularity of solar power, accurate prediction is crucial for the stable operation of power systems. Numerical weather predictions are a key factor affecting the accuracy of solar power forecasting. Therefore, we consider using machine learning methods to post-process numerical weather predictions to improve prediction accuracy. In this paper, we compare several common machine learning methods, such as linear regression, support vector machines, decision trees, and neural networks, in their performance when dealing with numerical weather predictions.<br><br>Method:<br><br>We selected four common machine learning methods for comparison. For each method, we use historical data and numerical weather predictions for training and use another data set for testing. We evaluate the prediction accuracy of each method and their performance under different conditions.<br><br>Results:<br><br>The experimental results show that different machine learning methods perform differently when dealing with numerical weather predictions. Under some conditions, certain methods such as support vector machines and neural networks can significantly improve prediction accuracy.<br><br>Conclusion:<br><br>Through the comparison of several machine learning methods, we found that under certain conditions, certain machine learning algorithms can significantly improve the accuracy of solar power forecasting. This provides possibilities for further optimizing prediction methods. Future work will include researching more machine learning methods and trying to improve prediction effects by adjusting parameters. |
|---|---|
| SparkDesk | Title: A Comparative Study on Machine Learning Methods-Based Numerical Weather Prediction (NWP) Post-Processing for One-Day-Ahead Photovoltaic Power Forecasting<br><br>Abstract:<br><br>Photovoltaic power forecasting is a critical component of the renewable energy industry, as it enables accurate planning and management of solar power generation. One-day-ahead photovoltaic power forecasting plays a crucial role in this process by providing information about future solar irradiance to optimize the operation of solar power plants. In recent years, machine learning (ML) methods have been increasingly employed for NWP post-processing to improve the accuracy and reliability of PV power forecasts. This paper presents a comparative study of ML-based NWP post-processing techniques for one-day-ahead photovoltaic power forecasting. The study focuses on four commonly used ML algorithms: Random Forest, Support Vector Machines (SVM), Neural Networks, and Recurrent Neural Networks (RNNs). The performance of these algorithms was evaluated using a large dataset of historical PV power forecasts and real-world data. The results demonstrate that RNNs outperformed other ML algorithms in terms of predicting PV power with high accuracy and stability. Additionally, the study highlights the importance of choosing the right ML algorithm for specific applications and the need for further research to optimize the performance of NWP post-processing for PV power forecasting.<br><br>Introduction<br><br>1.1 Background<br><br>1.2 Objectives<br><br>1.3 Methodology<br><br>Literature Review<br><br>2.1 Overview of NWP post-processing techniques<br><br>2.2 ML algorithms for NWP post-processing<br><br>2.3 Comparison of ML algorithms for PV power forecasting |

| | Data Collection and Preprocessing |
| --- | --- |
| | 3.1 Data sources |
| | 3.2 Data preprocessing techniques |
| | Results and Analysis |
| | 4.1 Performance evaluation metrics |
| | 4.2 Comparison of ML algorithms in terms of prediction accuracy and stability |
| | 4.3 Case studies demonstrating the application of ML algorithms in PV power forecasting |
| | Discussion |
| | 5.1 Implications for the renewable energy industry |
| | 5.2 Limitations and challenges of using ML algorithms for NWP post-processing |
| | 5.3 Future research directions |
| | Conclusion |
| | 6.1 Summary of findings |
| | 6.2 Final thoughts on the use of ML algorithms in NWP post-processing for PV power forecasting |
| LLaMA-13B | Reading the document, I found the following. Please revise them. |
| | I didn't find here. |
| | To be honest, I am not familiar with the subject. |
| | "The second is that I found strange statements like 'and consequently it is difficult to utilize post-processed dataset to trigger real-time control.' " Doesn't it the case that prespech processing with some time-out? Can you explain why not using more suitable PW? |
| | https://arxiv.org/pdf/1908.01948.pdf In this paper modelling of Lorenz-96 system is used to produce a set of controlled Ensemble for single prediction. Can be used. Let me know if any questions. |

From above experimental results, it is observed that all the compared six LLMs can generate corresponding paper paragraphs for the given prompts. Next, this paper will evaluate through ChatGPT and human experts respectively.

**4.2 ChatGPT Assistant Evaluation**

Mature LLMs like ChatGPT and GPT-4 are commonly used to evaluate the performance of newly-developed LLMs [37]. This paper used the prompt templates given in literature [38,39] to evaluate the performance of the compared six LLMs. The prompt template is shown in Table 4.

Table 4 Prompts for ChatGPT assistant evaluation (adopted from literature [38])

| [System] |
| --- |
| We would like to request your feedback on the performance of six AI assistants in response to the user question displayed above. |
| Please rate the helpfulness, relevance, accuracy, level of details of their responses. Each assistant receives an overall score on a scale of 0 to 100, where a higher score indicates better overall performance. |
| Please first output a single line containing only six values indicating the scores for Assistant 1, 2, 3, 4, 5, 6, respectively. |

```
The six scores are separated by a space. In the subsequent line, please provide a comprehensive explanation of your evaluation, avoiding
any potential bias and ensuring that the order in which the responses were presented does not affect your judgment.
[Question]
{Q}
[The Start of Assistant 1's Answer]
{R1}
[The End of Assistant 1's Answer]
[The Start of Assistant 2's Answer]
{R2}
[The End of Assistant 2's Answer]
[The Start of Assistant 3's Answer]
{R3}
[The End of Assistant 3's Answer]
[The Start of Assistant 4's Answer]
{R4}
[The End of Assistant 4's Answer]
[The Start of Assistant 5's Answer]
{R5}
[The End of Assistant 5's Answer]
[The Start of Assistant 6's Answer]
{R6}
[The End of Assistant 6's Answer]
```

Due to the difficulty and expensiveness of using GPT-4, this paper uses ChatGPT to give the remarks of the above six LLMs. Four aspects including helpfulness, relevance, accuracy and level of details, are considered during evaluation. All LLMs are evaluated five times and the average score is used as the final score of each LLM. The score of six LLMs for wind energy power generation and solar power generation are shown in Table 5 and Table 6 respectively. The detailed remarks and the reasons why ChatGPT give such scores are given in Table 21-Table 30 of Appendix.

Table 5 ChatGPT's evaluation for wind energy-related paper generation

|         | **HouYi** | Claude | ChatGPT | ERNIE Bot | SparkDesk | LLaMA-13B |
|---------|-----------|--------|---------|-----------|-----------|-----------|
| Grade 1 | **95**    | 92     | 90      | 88        | 85        | 80        |
| Grade 2 | **95**    | 90     | 92      | 88        | 92        | 85        |
| Grade 3 | **95**    | 90     | 92      | 88        | 85        | 80        |
| Grade 4 | **90**    | 85     | **90**  | 80        | 85        | 70        |
| Grade 5 | 90        | 85     | 90      | 85        | **92**    | 80        |
| Mean    | **93**    | 88.4   | 90.8    | 85.8      | 87.8      | 79        |

Table 6 ChatGPT's evaluation for solar energy-related paper generation

|         | **HouYi** | Claude | ChatGPT | ERNIE Bot | SparkDesk | LLaMA-13B |
|---------|-----------|--------|---------|-----------|-----------|-----------|
| Grade 1 | **97**    | 95     | 92      | 88        | 80        | 70        |
| Grade 2 | **94**    | 85     | 90      | 91        | 88        | 75        |
| Grade 3 | **96** | 92 | 95 | 90 | 94 | 70 |
| Grade 4 | **98** | 92 | 95 | 88 | 86 | 70 |
| Grade 5 | 92 | 88 | **96** | 85 | 90 | 72 |
| Mean | **95.4** | 90.4 | 93.6 | 88.4 | 87.6 | 71.4 |

It is observed that the scores of the proposed HouYi model is the highest among all the six LLMs in both wind energy paper generation and solar energy paper generation. Comparing the obtained scores of six LLMs, the ability to generate academic papers on renewable energy of HouYi model is comparable to ChatGPT, slightly outperforms Claude, ERNIE Bot and SparkDesk, and significantly outperforms open-source LLaMA-13B model. This verifies the superiorities of the developed HouYi model.

**4.3 Human experts-based analytical hierarchy process for performance evaluation**

To further verify the superiorities of the proposed HouYi model, five human experts of renewable energy field were invited to evaluate the above six LLMs. Six metrics including helpfulness, relevance, accuracy, level of details, academic standard and experimental design, were used to evaluate the above six LLMs. Human experts evaluated the performance of each LLM for generating wind energy paper and solar energy paper respectively. The evaluation results for evaluating wind energy paper generation is given in Table 7-Table 12, and the evaluation results for solar energy paper generation is given in Table 13-Table 18.

Table 7 Human experts' evaluation for helpfulness metrics of wind energy paper generation

| | **HouYi** | Claude | ChatGPT | ERNIE Bot | SparkDesk | LLaMA-13B |
|---|---|---|---|---|---|---|
| Expert 1 | 90 | 92 | 91 | 90 | 89 | 70 |
| Expert 2 | 92 | 93 | 90 | 88 | 87 | 72 |
| Expert 3 | 94 | 92 | 92 | 89 | 93 | 68 |
| Expert 4 | 92 | 91 | 91 | 92 | 90 | 75 |
| Expert 5 | 91 | 90 | 93 | 91 | 92 | 72 |
| Mean | **91.8** | 91.6 | 91.4 | 90 | 90.2 | 71.4 |

Table 8 Human experts' evaluation for relevance metrics of wind energy paper generation

| | **HouYi** | Claude | ChatGPT | ERNIE Bot | SparkDesk | LLaMA-13B |
|---|---|---|---|---|---|---|
| Expert 1 | 92 | 89 | 89 | 92 | 92 | 55 |
| Expert 2 | 93 | 92 | 91 | 91 | 90 | 52 |
| Expert 3 | 90 | 90 | 90 | 88 | 90 | 53 |
| Expert 4 | 89 | 93 | 92 | 89 | 88 | 58 |
| Expert 5 | 91 | 90 | 90 | 90 | 91 | 59 |
| Mean | **91** | 90.8 | 90.4 | 90 | 90.2 | 55.4 |

Table 9 Human experts' evaluation for accuracy metrics of wind energy paper generation

|  | **HouYi** | Claude | ChatGPT | ERNIE Bot | SparkDesk | LLaMA-13B |
|---|---|---|---|---|---|---|
| Expert 1 | 88 | 92 | 91 | 90 | 89 | 62 |
| Expert 2 | 88 | 91 | 90 | 89 | 90 | 64 |
| Expert 3 | 89 | 90 | 92 | 91 | 90 | 63 |
| Expert 4 | 88 | 92 | 89 | 89 | 91 | 65 |
| Expert 5 | 87 | 90 | 88 | 88 | 89 | 66 |
| Mean | 88 | **91** | 90 | 89.4 | 89.8 | 64 |

Table 10 Human experts' evaluation for level of details metrics of wind energy paper generation

|  | **HouYi** | Claude | ChatGPT | ERNIE Bot | SparkDesk | LLaMA-13B |
|---|---|---|---|---|---|---|
| Expert 1 | 98 | 89 | 88 | 92 | 88 | 70 |
| Expert 2 | 99 | 90 | 90 | 93 | 90 | 72 |
| Expert 3 | 97 | 85 | 86 | 90 | 85 | 73 |
| Expert 4 | 96 | 88 | 88 | 91 | 89 | 75 |
| Expert 5 | 97 | 85 | 86 | 92 | 87 | 76 |
| Mean | **97.4** | 87.4 | 87.6 | 91.6 | 87.8 | 73.2 |

Table 11 Human experts' evaluation for academic standard metrics of wind energy paper generation

|  | **HouYi** | Claude | ChatGPT | ERNIE Bot | SparkDesk | LLaMA-13B |
|---|---|---|---|---|---|---|
| Expert 1 | 97 | 92 | 93 | 96 | 95 | 62 |
| Expert 2 | 96 | 91 | 92 | 94 | 92 | 63 |
| Expert 3 | 95 | 93 | 95 | 93 | 94 | 65 |
| Expert 4 | 96 | 94 | 94 | 92 | 93 | 66 |
| Expert 5 | 98 | 90 | 91 | 91 | 91 | 68 |
| Mean | **96.4** | 92 | 93 | 93.2 | 93 | 64.8 |

Table 12 Human experts' evaluation for experimental design metrics of wind energy paper generation

|  | **HouYi** | Claude | ChatGPT | ERNIE Bot | SparkDesk | LLaMA-13B |
|---|---|---|---|---|---|---|
| Expert 1 | 98 | 88 | 89 | 90 | 87 | 58 |
| Expert 2 | 96 | 85 | 86 | 91 | 84 | 59 |
| Expert 3 | 99 | 86 | 85 | 90 | 84 | 55 |
| Expert 4 | 96 | 83 | 82 | 91 | 85 | 53 |
| Expert 5 | 95 | 84 | 82 | 89 | 86 | 52 |
| Mean | **96.8** | 85.2 | 84.8 | 90.2 | 85.2 | 55.4 |

Table 13 Human experts' evaluation for helpfulness metrics of solar energy paper generation

|  | **HouYi** | Claude | ChatGPT | ERNIE Bot | SparkDesk | LLaMA-13B |
|---|---|---|---|---|---|---|
| Expert 1 | 90 | 90 | 89 | 88 | 87 | 40 |
| Expert 2 | 91 | 88 | 88 | 89 | 86 | 45 |
| Expert 3 | 93 | 92 | 91 | 90 | 85 | 48 |
| Expert 4 | 92 | 91 | 90 | 87 | 84 | 42 |
| Expert 5 | 91 | 89 | 88 | 88 | 85 | 44 |
| Mean | **91.4** | 90 | 89.2 | 88.4 | 85.4 | 43.8 |

Table 14 Human experts' evaluation for relevance metrics of solar energy paper generation

|  | **HouYi** | Claude | ChatGPT | ERNIE Bot | SparkDesk | LLaMA-13B |
|---|---|---|---|---|---|---|
| Expert 1 | 96 | 93 | 92 | 91 | 92 | 40 |
| Expert 2 | 95 | 94 | 94 | 92 | 94 | 42 |
| Expert 3 | 98 | 95 | 95 | 90 | 88 | 48 |
| Expert 4 | 97 | 94 | 92 | 89 | 90 | 45 |
| Expert 5 | 95 | 95 | 95 | 93 | 92 | 50 |
| Mean | **96.2** | 94.2 | 93.6 | 91 | 91.2 | 45 |

Table 15 Human experts' evaluation for accuracy metrics of solar energy paper generation

|  | **HouYi** | Claude | ChatGPT | ERNIE Bot | SparkDesk | LLaMA-13B |
|---|---|---|---|---|---|---|
| Expert 1 | 92 | 91 | 90 | 88 | 90 | 55 |
| Expert 2 | 93 | 90 | 88 | 87 | 91 | 58 |
| Expert 3 | 90 | 91 | 89 | 89 | 88 | 50 |

|         | HouYi | Claude | ChatGPT | ERNIE Bot | SparkDesk | LLaMA-13B |
|---------|-------|--------|---------|-----------|-----------|-----------|
| Expert 4 | 91 | 89 | 92 | 90 | 87 | 52 |
| Expert 5 | 92 | 90 | 90 | 92 | 88 | 60 |
| Mean    | **91.6** | 90.2 | 89.8 | 89.2 | 88.8 | 55 |

Table 16 Human experts' evaluation for level of details metrics of solar energy paper generation

|         | HouYi | Claude | ChatGPT | ERNIE Bot | SparkDesk | LLaMA-13B |
|---------|-------|--------|---------|-----------|-----------|-----------|
| Expert 1 | 98 | 85 | 85 | 90 | 80 | 52 |
| Expert 2 | 95 | 83 | 86 | 91 | 82 | 55 |
| Expert 3 | 96 | 80 | 84 | 92 | 83 | 58 |
| Expert 4 | 97 | 85 | 83 | 91 | 85 | 53 |
| Expert 5 | 99 | 88 | 80 | 90 | 80 | 50 |
| Mean    | **97** | 84.2 | 83.6 | 90.8 | 82 | 53.6 |

Table 17 Human experts' evaluation for academic standard metrics of solar energy paper generation

|         | HouYi | Claude | ChatGPT | ERNIE Bot | SparkDesk | LLaMA-13B |
|---------|-------|--------|---------|-----------|-----------|-----------|
| Expert 1 | 95 | 92 | 93 | 95 | 92 | 50 |
| Expert 2 | 93 | 95 | 92 | 93 | 93 | 47 |
| Expert 3 | 94 | 93 | 93 | 92 | 92 | 52 |
| Expert 4 | 96 | 94 | 95 | 94 | 91 | 55 |
| Expert 5 | 97 | 95 | 95 | 92 | 90 | 53 |
| Mean    | **95** | 93.8 | 93.6 | 93.2 | 91.6 | 51.4 |

Table 18 Human experts' evaluation for experimental design metrics of solar energy paper generation

|         | HouYi | Claude | ChatGPT | ERNIE Bot | SparkDesk | LLaMA-13B |
|---------|-------|--------|---------|-----------|-----------|-----------|
| Expert 1 | 93 | 80 | 82 | 88 | 82 | 55 |
| Expert 2 | 92 | 82 | 83 | 87 | 83 | 58 |
| Expert 3 | 94 | 81 | 85 | 89 | 84 | 56 |
| Expert 4 | 95 | 83 | 84 | 88 | 82 | 53 |
| Expert 5 | 96 | 85 | 86 | 87 | 84 | 57 |
| Mean    | **94** | 82.2 | 84 | 87.8 | 83 | 55.8 |

It is observed from Table 7-Table 18 that the proposed HouYi model has better performance than conventional LLMs in almost all metrics for wind and solar energy field. To further give a comprehensive score, Analytical Hierarchy Process (AHP) [40] was used to give the weights of the six metrics including helpfulness, relevance, accuracy, level of details, academic standard and experimental design. AHP combines qualitative and quantitative analysis and is especially suitable to problems that are difficult to fully quantify. This paper used AHP to give the weights of the six metrics and thus obtain a comprehensive score for six LLMs. For details about AHP, please refer to literature [40].

According to experts' suggestions, the judgement matrix of the six metrics is given in Table 19. Through AHP, the maximum eigenvalue $\lambda_{max}$ is 6.5232 and the Consistency Ratio (CR) equals 0.0831 and satisfies the criterion $CR < 0.1$. Thus, the consistency test of judgement matrix is passed and the weight of six metrics can be given in the last column of Table 19.

Table 19 Human experts' evaluation for helpfulness metrics of wind energy paper generation

|  | Helpfulness | Relevance | Accuracy | Level of Details | Academic Standard | Experimental Details | weight |
|---|---|---|---|---|---|---|---|
| Helpfulness | 1 | 1/3 | 1 | 2 | 1/3 | 1/3 | 0.0986 |
| Relevance | 3 | 1 | 1 | 4 | 3 | 1/2 | 0.2568 |
| Accuracy | 1 | 1 | 1 | 1 | 1 | 1 | 0.1533 |
| Level of Details | 1/2 | 1/4 | 1 | 1 | 1/2 | 1/2 | 0.0851 |
| Academic Standard | 3 | 1/3 | 1 | 2 | 1 | 1 | 0.1703 |
| Experimental Details | 4 | 2 | 1 | 2 | 1 | 1 | 0.2359 |

According to the weights of six metrics in the last column of Table 19, the comprehensive score of each LLM can be computed as follows:

$$score = \sum_{i=1}^{6} w_i s_i \quad (5)$$

where $w_i$ is the weight of six metrics, $s_i$ is the scores of each LLMs in terms of metrics $i$, and $score$ is the comprehensive score of each LLM. The scores for the above six LLMs are given in Table 20.

Table 20 Comprehensive scores for six LLMs

|  | **HouYi** | Claude | ChatGPT | ERNIE Bot | SparkDesk | LLaMA-13B |
|---|---|---|---|---|---|---|
| Wind energy | **93.45** | 89.50 | 89.32 | 90.64 | 89.23 | 61.41 |
| Solar energy | **94.37** | 89.42 | 89.47 | 90.07 | 87.61 | 50.78 |

It is observed from Table 20 that the proposed HouYi model obtains a little higher score than four current close-source LLMs including Claude, ChatGPT, ERNINE Bot and SparkDesk, and obtains significantly higher score than open-source LLaMA-13B. This verifies the superiorities of the proposed HouYi model. The proposed HouYi model is promising to help renewable energy researchers.

## 5. Conclusion and future work

Developing renewable energy is important for protecting environment and realizing carbon neutrality goal. With the great success of Large Language Models (LLMs) like ChatGPT, LLMs are revolutionizing people's life and helping researchers with academic writing. However, current LLMs are mostly general LLMs and are not designed for a specific domain. When applying to a specific domain, general LLMs cannot obtain the optimal performance. Currently, there has not been a specially designed LLM for renewable energy. Meanwhile, there has not been any dataset of renewable energy for training LLMs.

To address these problems, this paper published the first open-source Renewable Energy Academic Paper (REAP) dataset for non-commercial LLM research of renewable energy. REAP dataset contains 1,168, 970 academic literatures from Web of Science, with the title and abstract being the input and output respectively. Based on REAP dataset, HouYi ("后羿" in Chinese) model, the first LLM for renewable energy, is developed through finetuning general LLMs. This paper used two methods to evaluate LLMs. The first one is ChatGPT's automatic evaluation and the second one is the human experts-based Analytical Hierarchy Process (AHP) method. To the best of our knowledge, this is the first time that AHP has been used in LLM evaluation.

Experiments show that HouYi demonstrated powerful academic paper paragraph generation ability in renewable energy field. HouYi obtains the highest score in wind and solar energy paper generation. Its ability to generate academic papers on renewable energy is comparable to ChatGPT, slightly outperforms Claude, ERNIE Bot and SparkDesk, and significantly outperforms open-source LLaMA-13B model. The web demo is available at https://renewableenergyllm.cpolar.cn/.The code and dataset are available at https://github.com/mingliangbai/HouYi and will be public immediately after acceptance.

**Acknowledgement**

This work is supported by National Science Technology Major Project of China No. 2017-I-0007-0008. The authors would like to thank the anonymous reviewers for their valuable suggestions to refine this work.

**Appendix**

**Appendix 1 ChatGPT's detailed evaluation for wind energy text generation task**

Table 21 ChatGPT's remark 1 for wind energy text generation task

> The overall scores for the six AI assistants are as follows: 95 92 90 88 85 80.
>
> Now, let's provide a comprehensive evaluation of their responses in the order they were presented:
>
> Assistant 1:
>
> Helpfulness: 95
>
> Relevance: 90
>
> Accuracy: 92
>
> Level of detail: 95
>
> Assistant 1's response is highly helpful, relevant, and accurate. It provides a detailed explanation of uncertainty estimation for wind energy conversion through probabilistic wind turbine power curve modelling. The response covers the abstract, introduction, methodology, case study, and conclusion, giving a comprehensive overview of the topic. The information is well-structured and organized, making it easy to understand and follow the proposed methodology. The case study adds practical value by demonstrating the application of the modelling technique. Overall, Assistant 1's response is excellent.
>
> Assistant 2:
>
> Helpfulness: 92
>
> Relevance: 90
>
> Accuracy: 85
>
> Level of detail: 90
>
> Assistant 2's response is also helpful and relevant, providing an outline for the paper on uncertainty estimation for wind energy conversion. The outline covers the major sections of the paper, including introduction, deterministic vs. probabilistic power curve models, sources of uncertainty, modelling methodologies, case studies, applications, and future directions. While the outline is well-structured and provides a good starting point, the response lacks the level of detail and accuracy exhibited by Assistant 1. Some sections could benefit from further elaboration and explanation. Nonetheless, Assistant 2's response is valuable in providing a high-level overview of the topic.
>
> Assistant 3:
>
> Helpfulness: 90
>
> Relevance: 95
>
> Accuracy: 90
>
> Level of detail: 90
>
> Assistant 3's response is helpful, relevant, and accurate, providing a structured outline for a paper on uncertainty estimation for wind energy conversion. The outline covers the introduction, sources of uncertainty, probabilistic modelling of power curves, methodology, case study, implications and applications, challenges, and future directions. The response demonstrates a good understanding of the topic and highlights the need for uncertainty estimation in wind energy projects. The level of detail is appropriate, and the outline provides a solid foundation for developing the paper further.

Assistant 4:

Helpfulness: 88

Relevance: 85

Accuracy: 92

Level of detail: 90

Assistant 4's response is helpful and accurate, presenting an overview of uncertainty estimation for wind energy conversion through probabilistic wind turbine power curve modelling. The response outlines the challenges associated with modelling wind turbine power curves and proposes a method that involves data collection, probability distribution selection, parameter estimation, and performance evaluation. The response provides a good understanding of the topic, but it lacks the same level of relevance and detail exhibited by previous assistants. Some sections could benefit from further elaboration and explanation.

Assistant 5:

Helpfulness: 85

Relevance: 90

Accuracy: 85

Level of detail: 80

Assistant 5's response is moderately helpful and relevant, providing a brief overview of uncertainty estimation for wind energy conversion through probabilistic wind turbine power curve modelling. The response highlights the sensitivity of wind turbines to various factors and presents the benefits of using probabilistic modelling. However, the information provided is relatively brief and lacks the same level of detail and accuracy compared to previous assistants. Some sections could benefit from further explanation and expansion.

Assistant 6:

Helpfulness: 80

Relevance: 80

Accuracy: 80

Level of detail: 80

Assistant 6's response is somewhat helpful and relevant, providing a title for the paper on uncertainty estimation for wind energy conversion. However, the response lacks the necessary level of detail and explanation to be considered highly valuable.

Table 22 ChatGPT's remark 2 for wind energy text generation task

The scores for the six AI assistants are as follows:

Assistant 1: 95

Assistant 2: 90

Assistant 3: 92

Assistant 4: 88

Assistant 5: 93

Assistant 6: 85

Now, let's provide a comprehensive evaluation of their performance, taking into account the helpfulness, relevance, accuracy, and level of details of their responses. Please note that the order in which the responses were presented does not influence the evaluation.

Assistant 1 provides a well-structured and detailed response. It covers all the necessary sections for a paper on uncertainty estimation for wind energy conversion by probabilistic wind turbine power curve modelling. The abstract, introduction, methodology, case study, and conclusion are all well-explained and relevant to the topic. The information is accurate, and the response provides a comprehensive overview of the subject. The level of detail is high, making it a helpful and informative answer. The response is concise and effectively covers all the key points.

Assistant 2 also provides a comprehensive response, offering an outline for the paper. It covers all the necessary sections and provides relevant information on deterministic and probabilistic wind turbine power curve models, sources of uncertainty, modelling methodologies, case studies, applications, and future directions. The information is accurate and provides a good starting point for further development of the paper. The level of detail is appropriate, and the response is helpful in guiding the structure and content of the paper.

Assistant 3 provides a well-organized response, presenting a title, abstract, introduction, sections on uncertainty, probabilistic modelling, methodology, case study, implications, challenges, and future directions. The information is relevant and accurate, covering the key aspects of uncertainty estimation for wind energy conversion. The level of detail is sufficient, providing a good overview of the topic. The response is helpful in understanding the main components of the paper and offers valuable insights for further exploration.

Assistant 4 offers a detailed response, discussing the importance of wind energy conversion, the challenges of modelling wind turbine power curves, and the benefits of probabilistic modelling. The response provides a step-by-step approach for constructing a probabilistic wind turbine power curve model and emphasizes the accuracy and stability of the proposed model. The information is accurate and relevant to the topic. The level of detail is high, making it a helpful and informative answer. The response effectively covers the main points and offers practical insights into uncertainty estimation for wind energy conversion.

Assistant 5 presents a well-structured response, starting with an abstract and then discussing the relevance of wind energy and the challenges associated with wind turbine power output variations. The response introduces a probabilistic wind turbine power curve modelling approach and highlights the benefits and challenges of this method. The information is accurate and provides a good understanding of the topic. The level of detail is appropriate, offering a comprehensive overview without being overly complex. The response is helpful in understanding the proposed approach and its potential benefits for wind energy conversion.

Assistant 6 provides a concise response, discussing the key points related to uncertainty estimation for wind energy conversion. The response highlights the importance of accurate power curve modelling and the challenges associated with wind turbine performance. While the information is accurate and relevant, the level of detail is relatively low compared to other responses. However, the response still offers a basic understanding of the topic and can serve as an introduction to uncertainty estimation for wind energy conversion.

Overall, all six AI assistants provide valuable insights and relevant information on uncertainty estimation for wind energy conversion by probabilistic wind turbine power curve modelling. They demonstrate a good understanding of the topic and offer helpful explanations.

Based on the evaluation criteria, Assistant 1 has the highest overall score of 95, followed closely by Assistants 3 and 5 with scores of 92 and 93, respectively. Assistants

Table 23 ChatGPT's remark 3 for wind energy text generation task

The scores for the six AI assistants are as follows: 95 90 92 88 85 80.

Now, let's provide a comprehensive evaluation of their performance without any bias and considering the order in which the responses were presented:

Assistant 1 (Score: 95):
Assistant 1 provides a detailed and well-structured response to the given topic. It starts with an abstract, introduction, methodology, a case study, and concludes with a summary. The response covers the background, methodology, and results of uncertainty estimation for wind energy conversion using probabilistic wind turbine power curve modelling. The case study adds practicality to the paper, showcasing the application of the method. The response demonstrates a strong understanding of the topic, and the explanation is clear and informative.

Assistant 2 (Score: 90):
Assistant 2 presents an outline for the paper on uncertainty estimation for wind energy conversion. The outline covers all the essential sections required in a research paper, including an introduction, literature review, methodology, case studies, and future directions. The response suggests different modelling techniques and their strengths and limitations. It also emphasizes the importance of understanding different sources of uncertainty and their impact on power output. The outline provides a solid foundation for further development of the paper.

Assistant 3 (Score: 92):
Assistant 3 provides a well-structured response with a clear title, abstract, introduction, methodology, case study, implications, challenges, and future directions. The response highlights the need for uncertainty estimation in wind energy conversion and explains the benefits of probabilistic modelling. The methodology section describes the steps involved in constructing probabilistic power curves, and the case study showcases the application of the method. The response demonstrates a good understanding of the topic and offers valuable insights.

Assistant 4 (Score: 88):
Assistant 4 presents a detailed explanation of the topic, covering the challenges in modelling wind turbine power curves and the advantages of using probabilistic methods. The response describes the steps involved in probabilistic modelling and provides a case study with the use of the Weibull distribution. The response demonstrates a good understanding of the topic and offers insights into the potential applications of the proposed method. However, the response could have been more structured and organized.

Assistant 5 (Score: 85):
Assistant 5 provides a concise response with a clear title and abstract. It introduces the concept of uncertainty estimation for wind energy conversion using probabilistic wind turbine power curve modelling. The response mentions the factors affecting wind turbine power output and describes the proposed probabilistic modelling approach. It also highlights the challenges and potential benefits of using probabilistic modelling. However, the response lacks detailed methodology and case study examples, which could have improved its overall performance.

Assistant 6 (Score: 80):

Assistant 6 provides a brief response that covers the basics of uncertainty estimation for wind energy conversion using probabilistic wind turbine power curve modelling. The response mentions the challenges in modelling wind turbine power curves and the benefits of probabilistic methods. However, it lacks detailed explanations and examples, which limits its usefulness. The response could have provided more information on the methodology and practical applications of the proposed approach.

Overall, Assistant 1 received the highest score (95) as it provided a comprehensive and well-structured response with detailed explanations and a practical case study. Assistant 2 (90) and Assistant 3 (92) also performed well by presenting clear outlines and covering essential aspects of the topic. Assistant 4 (88) offered a detailed explanation but could have been more organized. Assistant 5 (85) provided a concise response but lacked depth and examples. Assistant 6 (80) gave a brief overview but lacked sufficient detail and examples.

Table 24 ChatGPT's remark 4 for wind energy text generation task

Assistant 1: 90

Assistant 2: 85

Assistant 3: 90

Assistant 4: 80

Assistant 5: 85

Assistant 6: 70

Explanation:

Assistant 1 provides a detailed and comprehensive response to the question. It covers the different sections expected in a paper, including abstract, introduction, methodology, case study, and conclusion. The information is relevant and accurate, providing an overview of uncertainty estimation for wind energy conversion. The level of detail is sufficient, and the response receives a high score in all aspects.

Assistant 2 also provides a well-structured response with relevant information. It outlines the main sections of a paper and suggests possible content under each section. The response covers the topic comprehensively and highlights the benefits and limitations of probabilistic power curve modelling. The information is accurate and helpful, but it lacks some of the specific details provided by Assistant 1, resulting in a slightly lower score.

Assistant 3's response is similar to Assistant 1 in terms of structure and content. It covers the necessary sections, provides relevant information, and emphasizes the importance of uncertainty estimation in wind energy conversion. The response is accurate, helpful, and detailed, earning a high score.

Assistant 4 provides a concise summary of the topic, covering the main points such as the challenges in modelling wind turbine power curves and the use of probabilistic methods. While the response is accurate and provides valuable information, it lacks some of the specific details and sections present in the other responses. Therefore, it receives a slightly lower score.

Assistant 5 offers a well-structured response, outlining the main sections of a paper on uncertainty estimation for wind energy conversion. It provides relevant information about the proposed probabilistic modelling approach and its potential benefits. The response is accurate and helpful, but it lacks some of the specific details and examples provided by Assistant 1 and Assistant 3, resulting in a slightly lower score.

Assistant 6 provides a brief response that lacks the structure and comprehensive coverage of the topic. While it mentions uncertainty estimation and wind energy conversion, it does not provide any specific information or outline the main sections of a paper. The response lacks relevance, accuracy, and sufficient details, resulting in a lower score.

Overall, Assistant 1, Assistant 3, and Assistant 2 demonstrate the highest performance with their comprehensive, accurate, and detailed responses.

Table 25 ChatGPT's remark 5 for wind energy text generation task

The overall scores for the six AI assistants are as follows:

Assistant 1: 90

Assistant 2: 85

Assistant 3: 90

Assistant 4: 85

Assistant 5: 92

Assistant 6: 80

Now, let's provide a comprehensive evaluation of their performance.

Assistant 1's response (score: 90):

Assistant 1 provides a detailed and comprehensive answer to the question. The response includes an abstract, introduction, methodology, case study, and conclusion. The information is well-structured, covering the uncertainty estimation for wind energy conversion using probabilistic wind turbine power curve modelling. The response explains the background, method, and results of the model, as well as provides a case study to illustrate its application. The content is highly relevant, accurate, and provides a good level of detail.

Assistant 2's response (score: 85):

Assistant 2 presents a well-structured outline for a paper on uncertainty estimation for wind energy conversion. The response covers various aspects, including the introduction, deterministic vs. probabilistic models, sources of uncertainty, modelling methodologies, case studies, applications, and future directions. The outline is relevant and provides a good overview of the topic. However, it lacks the level of detail and explanation seen in Assistant 1's response.

Assistant 3's response (score: 90):

Assistant 3 provides a well-organized response, starting with an abstract and followed by an introduction, sections on uncertainty in wind energy conversion, probabilistic modelling of wind turbine power curves, methodology, a case study, implications, challenges, and future directions. The content is highly relevant and accurately addresses the topic. It covers the sources of uncertainty, benefits of probabilistic modelling, and includes a case study to support the discussion. The level of detail is appropriate, making the response informative and valuable.

Assistant 4's response (score: 85):

Assistant 4 offers a clear and concise response, discussing uncertainty estimation for wind energy conversion using probabilistic wind turbine power curve modelling. The response covers topics such as deterministic and probabilistic methods, data collection and pre-processing, construction of probabilistic power curves, uncertainty estimation techniques, a case study, and implications. The information provided is accurate and relevant. However, the response lacks some depth and could benefit from further elaboration on the methodology and case study.

Assistant 5's response (score: 92):

Assistant 5 delivers a well-structured response, starting with an abstract and proceeding with an introduction, sections on uncertainty in wind energy conversion, probabilistic modelling of wind turbine power curves, methodology, a case study, implications, and future directions. The content is highly relevant and accurate. It covers the challenges associated with wind energy conversion, the benefits of probabilistic modelling, and includes a case study. The response provides a good level of detail, making it informative and comprehensive.

Assistant 6's response (score: 80):

Assistant 6 provides a concise response, discussing uncertainty estimation for wind energy conversion using probabilistic wind turbine power curve modelling. The response covers the importance of accurately predicting wind power generation, the challenges of modelling wind turbine power curves, and the benefits of probabilistic methods. However, the response lacks depth and detail, and it could benefit from providing more specific information and examples to support the discussion.

In summary, all six AI assistants provide relevant information about uncertainty estimation for wind energy conversion through probabilistic wind turbine power curve modelling. While some responses offer more comprehensive and detailed explanations than others, overall, they provide valuable insights into the topic. Assistant 1, Assistant 3, and Assistant 5 stand out with high scores, as they provide well-structured and informative responses.

## Appendix 2 ChatGPT's detailed evaluation for solar energy text generation task

Table 26 ChatGPT's remark 1 for solar energy text generation task

97 95 92 88 80 70

Explanation:

To evaluate the performance of the six AI assistants, I considered several factors such as helpfulness, relevance, accuracy, and level of detail in their responses. Here is a breakdown of my evaluation for each assistant:

Assistant 1: 97

Assistant 1 provided a comprehensive and well-structured response. It covered all the necessary sections for a research paper, including abstract, introduction, methodology, results, conclusion, and future work. The information provided was relevant, accurate, and detailed, offering insights into machine learning methods-based numerical weather prediction post-processing for photovoltaic power forecasting.

Assistant 2: 95

Assistant 2 also provided a well-structured response, offering an outline for the paper. The outline covered all the essential sections and provided suggestions for content within each section. The response was relevant and helpful in guiding the writing process for the paper.

Assistant 3: 92

Assistant 3 presented a detailed outline for the research paper, including all the necessary sections. The outline provided a clear structure and direction for the paper. However, it lacked some specific details and insights compared to the previous assistants.

Assistant 4: 88

Assistant 4 presented a reasonable response, outlining the necessary sections for the paper. While it covered the essential aspects, it lacked some depth and detail compared to the previous assistants. The response could have provided more specific information on the machine learning methods and their application to numerical weather prediction post-processing.

Assistant 5: 80

Assistant 5 provided a reasonable response with a well-structured outline. However, the outline lacked some specific details and insights into the topic. It could have included more information on the machine learning algorithms and their performance evaluation.

Assistant 6: 70

Assistant 6's response was not directly related to the given question and did not provide any relevant information or insights for the research paper. It appeared to be a random selection of text and did not contribute to the evaluation of the AI assistant's performance.

Overall, Assistant 1 demonstrated the highest level of performance with a comprehensive and detailed response, covering all the necessary aspects of the research paper. Assistant 2 and Assistant 3 also provided valuable insights and guidance. Assistant 4 had a reasonable response, while Assistant 5 lacked some depth. Assistant 6's response was unrelated and not helpful for the given question.

Table 27 ChatGPT's remark 2 for solar energy text generation task

94 85 90 91 88 75

Evaluation:

In this evaluation, I will assess the helpfulness, relevance, accuracy, and level of detail of the responses provided by the six AI assistants. I will provide an explanation for each aspect individually, ensuring a fair and unbiased evaluation.

Helpfulness:

Assistant 1: 94

Assistant 2: 85

Assistant 3: 90

Assistant 4: 91

Assistant 5: 88

Assistant 6: 75

The helpfulness score reflects the extent to which the responses provide relevant and valuable information to address the given question. Assistant 1 received the highest score as it directly answered the question and provided a comprehensive paper on the topic. Assistant 6 received the lowest score due to its lack of relevant information and its suggestion to refer to external sources.

Relevance:

Assistant 1: 94

Assistant 2: 85

Assistant 3: 90

Assistant 4: 91

Assistant 5: 88

Assistant 6: 75

The relevance score indicates how well the responses align with the given question. Assistant 1 received the highest score as it directly addressed the topic and provided relevant information throughout the response. Assistant 6 received the lowest score as it failed to address the question adequately and provided unrelated suggestions.

Accuracy:

Assistant 1: 90

Assistant 2: 85

Assistant 3: 91

Assistant 4: 91

Assistant 5: 88

Assistant 6: 75

The accuracy score reflects the correctness and precision of the information presented in the responses. Assistant 3, Assistant 4, and Assistant 5 received the highest scores as their responses provided accurate information about machine learning methods and their application to numerical weather prediction for photovoltaic power forecasting. Assistant 6 received the lowest score due to its inaccuracies and the suggestion to refer to an unrelated document.

Level of Detail:

Assistant 1: 90

Assistant 2: 88

Assistant 3: 91

Assistant 4: 91

Assistant 5: 90

Assistant 6: 75

The level of detail score assesses the extent to which the responses provide in-depth and comprehensive information. Assistant 3, Assistant 4, and Assistant 5 received the highest scores as their responses included detailed sections covering various aspects of the topic. Assistant 6 received the lowest score as its response lacked depth and sufficient details to address the question adequately.

In conclusion, the scores for the six AI assistants are as follows:

Assistant 1: 94 94 90 90

Assistant 2: 85 85 85 88

Assistant 3: 90 90 91 91

Assistant 4: 91 91 91 91

Assistant 5: 88 88 88 90

Assistant 6: 75 75 75 75

Assistant 1 demonstrated the highest overall performance, providing helpful, relevant, accurate, and detailed information that directly addressed the given question. Assistant 4 also performed well, with high scores in all aspects. Assistants 2, 3, and 5 showed reasonable performance, while Assistant 6 displayed the lowest overall performance, lacking relevance, accuracy, and detail.

Table 28 ChatGPT's remark 3 for solar energy text generation task

96 92 95 90 94 70

Explanation:

To evaluate the performance of the six AI assistants, we will assess their helpfulness, relevance, accuracy, and level of details in their responses. Here is the evaluation for each assistant:

Assistant 1: The response from Assistant 1 provides a comprehensive outline for the paper, including abstract, introduction, methodology, results, conclusion, and future work. It discusses the use of machine learning methods for improving PV power forecasting accuracy and presents a real-world case study. The response is highly relevant, accurate, and detailed, providing a clear structure for the paper. The overall performance of Assistant 1 is rated as 96.

Assistant 2: Assistant 2's response offers an outline for the paper, covering various sections related to the topic. It suggests discussing machine learning methods, NWP models, and their applications in PV power forecasting. While the response provides a good starting point, it lacks some details and specifics compared to Assistant 1. The overall performance of Assistant 2 is rated as 92.

Assistant 3: The response from Assistant 3 presents a structured outline for the paper, including an abstract, introduction, methodology, experimental results, discussion, conclusion, references, and an appendix. It covers the necessary components of the paper and offers a clear direction for the study. The response is relevant, accurate, and sufficiently detailed. The overall performance of Assistant 3 is rated as 95.

Assistant 4: Assistant 4's response provides a concise outline for the paper, including sections such as abstract, introduction, method, results, and conclusion. It briefly discusses the background, machine learning methods, methodology, and the experimental results. While the response lacks some details and specific information, it still offers a reasonable structure for the paper. The overall performance of Assistant 4 is rated as 90.

Assistant 5: The response from Assistant 5 presents a well-structured outline for the paper, covering sections such as abstract, introduction, literature review, data collection and pre-processing, results and analysis, and discussion. It highlights the importance of ML algorithms for NWP post-processing and discusses their performance in PV power forecasting. The response is highly relevant, accurate, and detailed. The overall performance of Assistant 5 is rated as 94.

Assistant 6: Assistant 6's response seems to have some confusion and lack of familiarity with the topic. It mentions that it didn't find certain information and expresses unfamiliarity with the subject. The response also provides a link to an unrelated paper. Due to the lack of relevant information and clarity, the overall performance of Assistant 6 is rated as 70.

In summary, based on the evaluation of their helpfulness, relevance, accuracy, and level of details, the overall scores for the six AI assistants are as follows: Assistant 1 (96), Assistant 2 (92), Assistant 3 (95), Assistant 4 (90), Assistant 5 (94), and Assistant 6 (70).

Table 29 ChatGPT's remark 4 for solar energy text generation task

98 92 95 88 86 70

Explanation:

Assistant 1: It provides a comprehensive answer that covers the title, abstract, introduction, methodology, results, conclusion, and future work of the paper. The response is detailed, relevant, and accurate, providing insights into the use of different machine learning methods for improving photovoltaic power forecasting. The level of detail is high, and the answer is helpful overall. (Score: 98)

Assistant 2: The response provides a structured outline for the paper, covering the essential sections and topics to be addressed. It offers guidance on the content and organization of the paper, highlighting key aspects to consider. While it lacks the depth of details in comparison to Assistant 1, the answer is relevant and helpful in terms of providing a framework for writing the paper. (Score: 92)

Assistant 3: Similar to Assistant 2, this response provides a well-structured outline for the paper, covering the major sections and elements to be included. It emphasizes data collection, methodology, experimental results, analysis, and discussions. While it lacks some of the specifics compared to Assistant 1, it still offers a helpful framework for the paper. (Score: 95)

Assistant 4: The response presents a title, abstract, introduction, method, results, and conclusion in a concise manner. While it provides an overview of the content, it lacks sufficient details and depth compared to Assistant 1. The information given is helpful but could benefit from further elaboration. (Score: 88)

Assistant 5: The response offers a structured outline of the paper, covering the essential sections and providing a good flow for the content. It mentions specific ML algorithms and highlights the importance of choosing the right algorithm for the task. While the level of detail is not as extensive as in Assistant 1, the response is still relevant and helpful. (Score: 86)

Assistant 6: The response seems to be incomplete and contains statements indicating a lack of familiarity with the subject. It suggests referring to an external document and provides a link without further explanation. The answer lacks relevance, accuracy, and helpfulness compared to the other assistants. (Score: 70)

Table 30 ChatGPT's remark 5 for solar energy text generation task

92 88 96 85 90 72

Overall, Assistant 3 received the highest score of 96, followed by Assistant 1 with a score of 92, Assistant 5 with a score of 90, Assistant 2 with a score of 88, Assistant 4 with a score of 85, and Assistant 6 with the lowest score of 72.

Assistant 1's response provided a detailed and comprehensive analysis of the comparative study on machine learning methods-based numerical weather prediction (NWP) post-processing for one-day-ahead photovoltaic power forecasting. It covered various aspects such as the abstract, introduction, methodology, results, conclusion, and future work. The response presented the study's objective, methodology, and results effectively, highlighting the effectiveness of deep learning methods (CNNs) and a combination of CNNs and SVMs for improving power forecasting accuracy. The response also discussed the significance of the study and future research directions. The overall helpfulness, relevance, accuracy, and level of detail were quite high.

Assistant 2's response provided an outline for the paper, covering key sections such as introduction, overview of PV power forecasting, numerical weather prediction (NWP) models, survey of machine learning methods, modelling methodology, discussions, conclusions, and future work. The outline presented a structured approach for the paper, discussing different aspects of the study, including the importance of ML methods for NWP post-processing, model selection, evaluation of performance, and future directions. Although it lacked specific details, the response was helpful, relevant, and provided a good starting point for developing the paper.

Assistant 3's response provided a well-structured outline for the paper, including sections such as abstract, introduction, methodology, experimental results and analysis, discussion, conclusion, references, and appendix. The outline covered all the essential components of the paper, presenting the study's background, objectives, methodology, experimental results, and future directions. It also emphasized the importance of ML methods for NWP post-processing and their application in enhancing one-day-ahead PV power forecasting. The response was highly detailed, relevant, accurate, and helpful.

> Assistant 4's response presented a clear structure for the paper, including sections such as abstract, introduction, method, results, and conclusion. It discussed the background, current status of solar power forecasting, machine learning methods' principles and applications, proposed approach, and experimental results. The response highlighted the importance of ML algorithms for improving prediction accuracy and provided insights into the performance of different methods under various conditions. Although it lacked some details, the response was overall helpful, relevant, and provided a good foundation for further development.
>
> Assistant 5's response provided a well-structured outline for the paper, covering sections such as abstract, introduction, literature review, data collection and pre-processing, results and analysis, and discussion. The outline discussed the importance of PV power forecasting, ML methods for NWP post-processing, data collection and pre-processing techniques, performance evaluation metrics, and comparison of ML algorithms. It also addressed the implications for the renewable energy industry, limitations, challenges, and future research directions. The response was comprehensive, relevant, accurate, and helpful.
>
> Assistant 6's response was not directly related to the given question. It mentioned not finding certain information and expressed unfamiliarity with the subject. The response did provide a link to a research paper related to controlled ensemble prediction, but it did not offer a comprehensive evaluation of the given question. Therefore, the response received a lower score in terms of helpfulness, relevance, accuracy, and level of detail.
>
> Overall, Assistant 3 provided the most comprehensive and detailed response, followed closely by Assistant 1, Assistant 5, and Assistant 2. Assistant 4 provided a good foundation, and Assistant 6's response was not relevant to the question.